\documentclass{article}
\usepackage{spconf,amsmath,epsfig}
\usepackage{url}

\title{Feature Selection on Sentinel-2 Multi-spectral Imagery for Efficient Tree Cover Estimation}
\name{Usman Nazir, Momin Uppal, Muhammad Tahir, and Zubair Khalid
\thanks{We acknowledge the support of the Higher Education Commission of Pakistan under grant GCF-521.}}
\address{Department of Electrical Engineering, Syed Babar Ali School of Science and Engineering\\
Lahore University of Management Sciences (LUMS), Lahore, Pakistan\\
\{usman.nazir, momin.uppal, tahir, zubair.khalid\}@lums.edu.pk}
\begin{document}
%
\maketitle
\begin{abstract}
This paper proposes a multi-spectral random forest classifier with suitable feature selection and masking for tree cover estimation in urban areas. The key feature of the proposed classifier is filtering out the built-up region using spectral indices followed by random forest classification on the remaining mask with carefully selected features.  Using Sentinel-2 satellite imagery, we evaluate the performance of the proposed technique on a specified area (approximately 82 acres) of Lahore University of Management Sciences (LUMS) and demonstrate that our method outperforms a conventional random forest classifier as well as state-of-the-art methods such as European Space Agency (ESA) WorldCover $10$m $2020$ product as well as a DeepLabv3 deep learning architecture.
\end{abstract}
\section{Introduction}
The presence of easily accessible multispectral satellite imagery has expanded the range of potential applications across diverse fields. An important example is automated detection of trees and green spaces that are significant contributors to ecosystem services such as air purification and carbon sequestration. Recent studies include  \cite{majasalmi2021representation} and \cite{grabska2019forest} for global monitoring of environment and forest cover using Sentinel-2 imagery. A Copernicus Sentinel-2B satellite, launched in 2017 provides 13 bands with spatial resolution from $10$ m to $60$ m. The high spatial and temporal resolution of data from this satellite is specifically designed for vegetation monitoring. For tree cover estimation, a broad range of methodologies have been presented in the literature, e.g., ~\cite{majasalmi2021representation, martinis2009towards, hansen2013high, negassa2020forest}. The authors in \cite{martinis2009towards} proposed a data fusion method of different spatial resolution satellite imagery for forest-type mapping. Forest cover change is mapped in \cite{hansen2013high} using spatial resolution of $25$ m to $30$ m. A spatio-temporal study on forest cover change using GIS and remote sensing techniques in Ethiopia valley is presented in \cite{negassa2020forest}. In \cite{majasalmi2021representation}, a simple tree cover (referred to as `forest cover') approach using three different land cover (LC) products is employed in Finland. Clearly, most of these approaches focus on forest mapping -- a gap exists in urban tree cover estimation in developing countries with low resolution imagery. 

In this paper, We propose a multi-spectral classifier (that uses a mixture of spectral bands \emph{and} indices) for tree cover estimation in urban areas. The key aspects of the proposed classifier include a masking stage for filtering out built-up areas, followed by a random forest classifier operating on appropriately selected features. For performance evaluation, we manually annotate $3768$ trees in Lahore, Pakistan\footnote{This dataset is being made publicly available at \url{https://city.lums.edu.pk/products/}}. We demonstrate that on account of suitable feature selection and masking mechanism, our proposed approach applied to low resolution imagery achieves a higher accuracy level compared to that obtained by the European Space Agency (ESA) WorldCover product~\cite{van2021esa} as well as a more computationally demanding deep learning architecture DeepLabv3 \cite{chen2017deeplab} applied on high-resolution imagery.

The subsequent sections of this paper are structured as follows: Section $2$ delves into a comprehensive analysis of the methodology, while Section $3$ showcases the evaluation results comparing our proposed methodology with state-of-the-art models. Finally, Section 4 concludes the paper.

\begin{figure*}[h]
  \centering
  \centerline{\includegraphics[width=18cm]{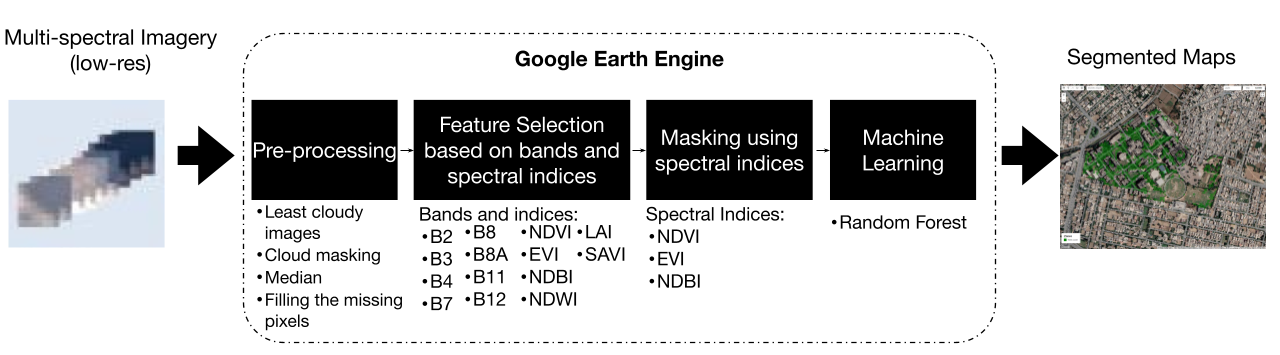}}
  \caption{Proposed methodology for tree cover estimation with feature selection of spectral bands and indices.}
  \label{fig:method}
\end{figure*}
\begin{table*}[h]
    \caption{Tree cover area is predicted using Sentinel-2 imagery and RF classifier by employing various feature selection techniques. The Lahore University of Management Sciences (LUMS) study region spans $82.165$ acres of area with $23$ acres area of actual tree cover.}
    \centering
    \label{tab:eval1}
    \scalebox{0.85}{
    \begin{tabular}{|c|c|c|c|c|c|c|}
    \hline
    \bf ROI & \bf Model &\bf Pred. area (acres) &\bf Masking & \bf Spectral indices &\bf Pixel-wise Test Accuracy (\%)& Kappa Score \\ \hline
    LUMS  & RF-spectral-bands &  29.5 & No & No  & 0.93 & 0.81 \\ 
    \hline 
    LUMS  & RF-spectral-indices  & 28  & No & Yes  & 0.95 & 0.88\\ 
    \hline 
    
    LUMS  & \bf Proposed &  25  & Yes & Yes &  \bf 0.99 & \bf 0.92\\ 
    
    \hline \hline
    
    LUMS  & ESA WorldCover Product~\cite{van2021esa} &  16 & - & -  & 0.74 & -\\ 
    LUMS & DeepLabv3~\cite{chen2017deeplab} &  28 & No & No & 0.80 & -\\ \hline     
    \end{tabular}}
\end{table*}
\begin{table*}[h]
    \caption{Pixel-wise accuracy and Kappa score of proposed model with different feature set on LUMS study region.}
    \centering
    \label{tab:eval2}
    \begin{tabular}{|c|c|c|}
    \hline
    \bf (RF + Masking +) \bf Features set &\bf Pixel-wise Test Accuracy (\%)& Kappa Score \\ \hline
    Eight multispectral bands + NDVI & 0.96 & 0.76 \\ \hline 
    
    
    Eight multispectral bands + NDVI + NDWI + NDBI + EVI & 0.97 & 0.80 \\ \hline
    Eight multispectral bands \& All spectral indices & \bf 0.99 & \bf 0.92 \\ \hline     
    \end{tabular}
\end{table*}
\section{Methodology}
The proposed methodology, illustrated in Fig.~\ref{fig:method}, consists of four stages. These include 1) Pre-processing, 2) Feature selection, 3) Masking, and finally 4) Random Forest Classification. The details of each stage are provided in the text below.
\subsection{Pre-processing} 
We divide the pre-processing of data into multiple steps. Initially, the images from a multi-spectral satellite containing less than $10\%$ cloud cover for the region of interest (LUMS) are passed through a cloud masking operation that removes cloud cover from these images. Next a median of these images is taken for each month. Finally multiple images are stacked together to generate a single combined image of the region of interest. 
\subsection{Feature selection}
For classification, we included  eight bands of Sentinel-2 imagery as the feature set. These include B2 (Blue), B3 (Green), B4 (Red), B7 (Red Edge 3), B8 (NIR), B8A (Red Edge 4), B11 (SWIR 1) and B12 (SWIR 2). In addition, We also chose six spectral indices in our feature set. These include the Normalized Difference Vegetation Index (NDVI)~\cite{rouse1974monitoring}, Enhanced Vegetation Index (EVI)~\cite{huete2002overview}, Normalized Difference Built-up Index (NDBI)~\cite{NDBI:online}, Normalized Difference Moisture Index (NDMI)~\cite{mcfeeters1996use}, Leaf Area Index (LAI)~\cite{boegh2002airborne} and Soil Adjusted Vegetation Index (SAVI)~\cite{huete1988soil}. In general, regions with tree cover typically exhibit high vegetation indices (EVI, NDVI), NDMI, LAI, and SAVI, while showing notably low values for NDBI. Some background about these indicies is given below.

\textbf{NDVI}: This index \cite{rouse1974monitoring} describes the difference between visible and near-infrared reflectance of vegetation cover and can be used to estimate the density of green on an area of land. This is computed from the the NIR and the Red bands measurements as follows
\begin{equation}
    \text{NDVI} = \frac{\text{NIR} - \text{Red}}{\text{NIR} + \text{Red}}
\end{equation}

\textbf{EVI and LAI:}
EVI~\cite{huete2002overview} is similar to NDVI and can be used to quantify greenness of vegetation. However, EVI corrects for some atmospheric conditions and canopy background noise and is more sensitive in areas with dense vegetation. It is computed as 
\begin{equation}
\text{EVI} = 2.5 \times \frac{\text{NIR} - \text{Red}}{\text{NIR} + 6\times \text{Red} - 7.5 \times \text{Blue} + 1}
\end{equation}
On the other hand, LAI~\cite{boegh2002airborne} is used to estimate crop growth and yield through the following empirical formula
\begin{equation}
\text{LAI} =3.618\times \text{EVI} - 0.118
\end{equation}

\textbf{SAVI}: This index \cite{huete1988soil} attempts to minimize soil brightness influences using a soil-brightness correction factor. This is often used in arid regions where vegetative cover is low, and it outputs values between $-1$ and $1$ through the following relationship
\begin{equation}
    \text{SAVI} = \frac{0.5 \times (\text{NIR}-\text{Red})}{\text{NIR}+\text{Red}+0.5}
\end{equation}

\begin{figure*}[h]
    \centering
    \scalebox{0.4}{
    \begin{tabular}{ccccc}
        \includegraphics[scale=0.3]{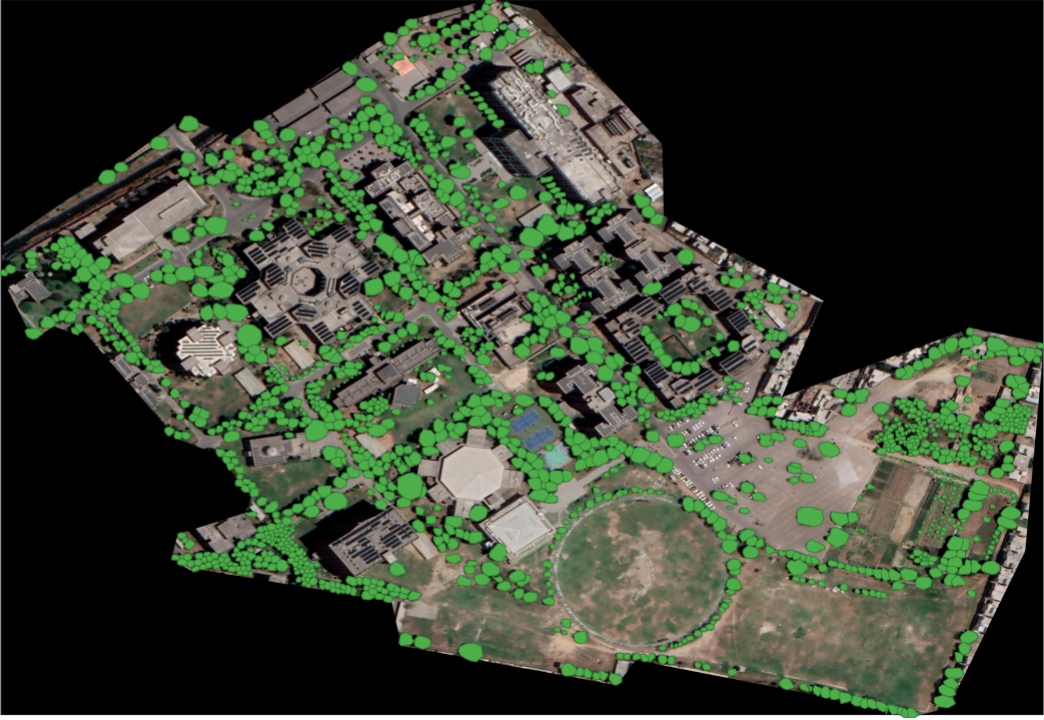} 
        & \includegraphics[scale=0.3]{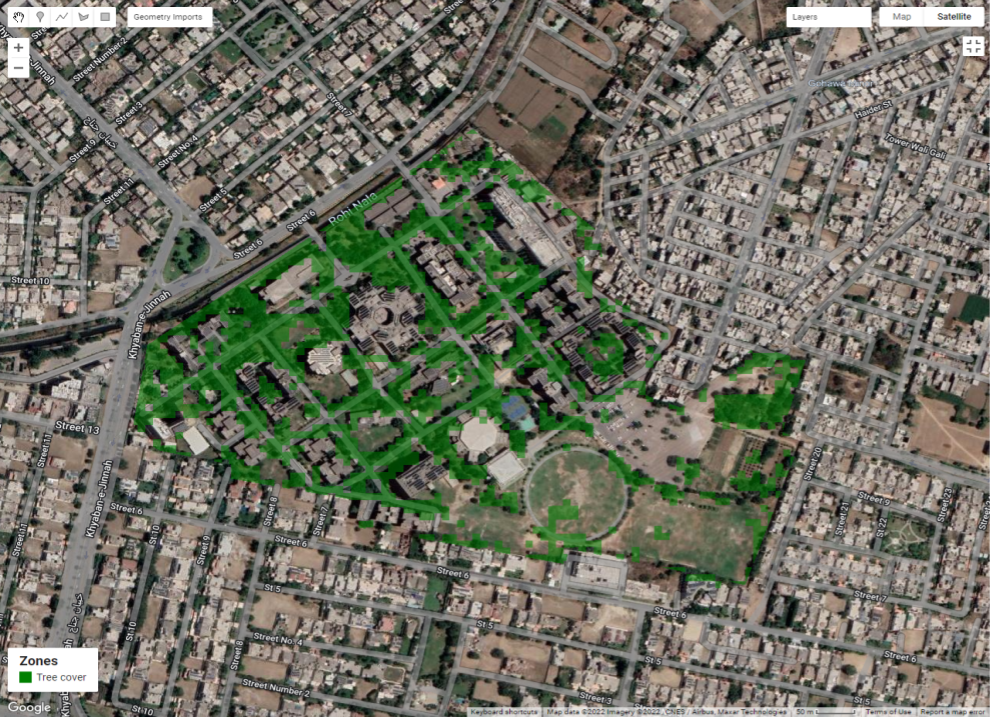} & \includegraphics[scale=0.3]{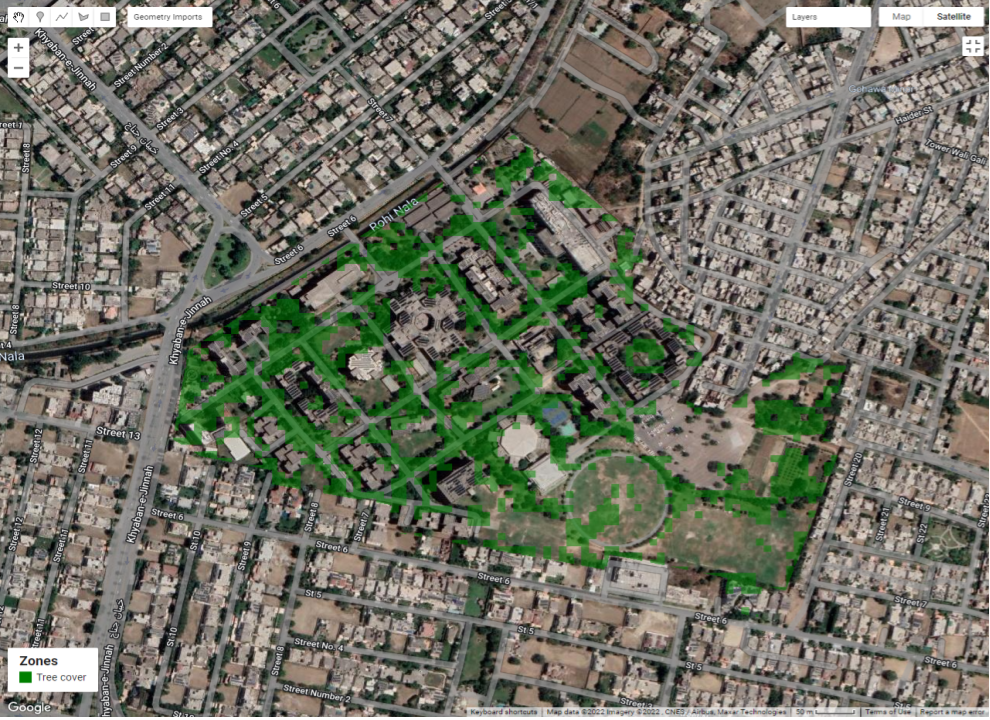}&\includegraphics[scale=0.3]{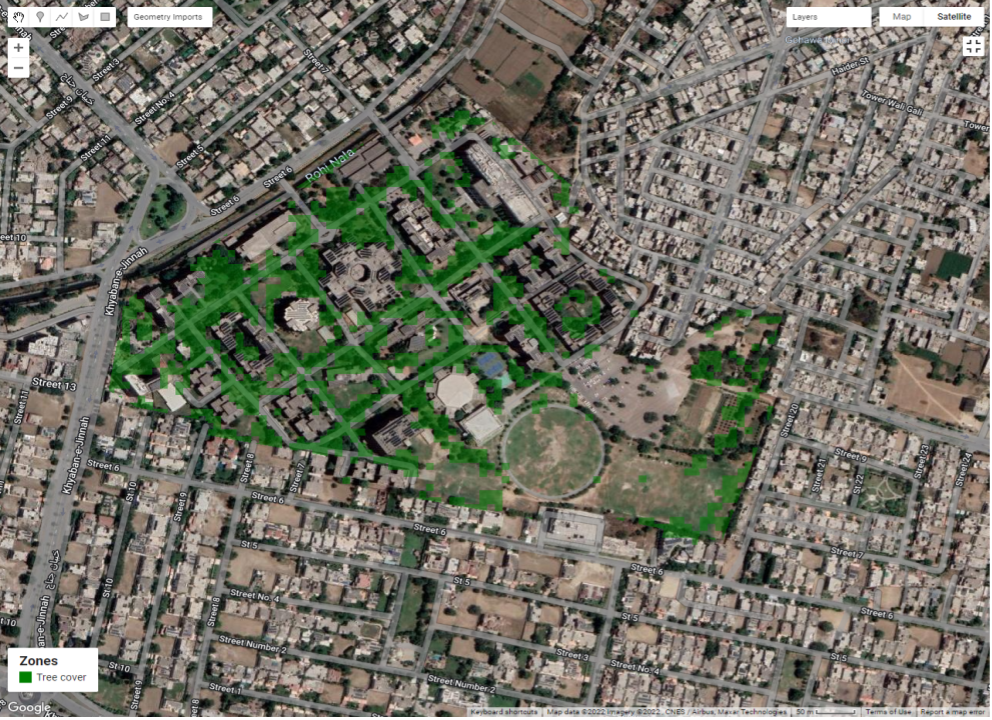} & \includegraphics[scale=0.3]{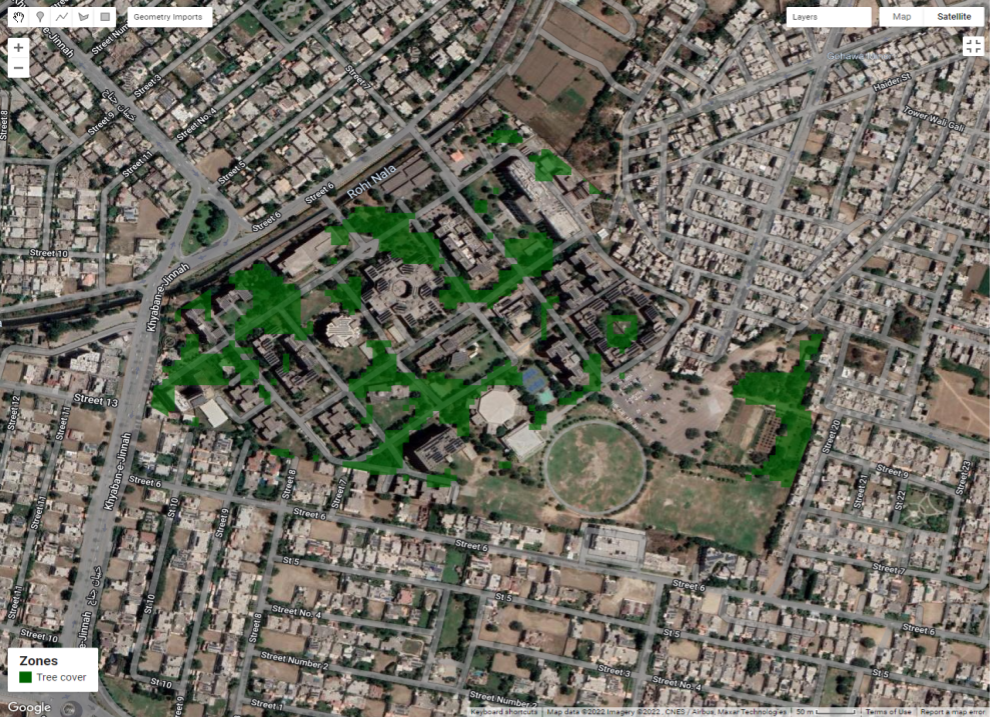} \\
        LUMS Ground Truth & RF (classical) & RF-Indices & Proposed & ESA WorldCover\\
    \end{tabular}}
    \caption{Qualitative results using feature selection on Sentinel-2 multi-spectral imagery for efficient tree cover estimation.}
    \label{fig:qualtitative}
\end{figure*}

\textbf{NDWI}: This is a satellite-derived index \cite{gao1996ndwi} from the NIR and the SWIR channels. The NDWI is used to monitor changes related to water content in water bodies as they strongly absorb light in visible to infrared electromagnetic spectrum. 
\begin{equation}
    \text{NDWI} = \frac{\text{NIR} - \text{SWIR1}}{\text{NIR} + \text{SWIR1}}
\end{equation}

\textbf{NDBI:} This index \cite{zha2003use} uses the NIR and SWIR bands to emphasize manufactured built-up areas. It aims to mitigate the effects of terrain illumination differences as well as atmospheric effects. 
\begin{equation}
\text{NDBI} = \frac{\text{SWIR} - \text{NIR2}}{\text{SWIR} + \text{NIR2}}    
\end{equation}

\subsection{Masking} 
Masking process involves the following two steps.

\noindent \emph{Applying the Vegetation Index}. The EVI or NDVI values are calculated for each pixel in the satellite imagery. These values indicate the presence and density of vegetation. In this case, a threshold of 0.2 is set, implying that any pixel with an EVI or NDVI value equal to or below 0.2 is considered non-vegetated or sparsely vegetated. Such regions are likely to include built-up areas, as they have less vegetation cover.

\noindent \emph{Utilizing the Built-up Index}. Simultaneously, the NDBI values are computed for each pixel. This index highlights the presence and extent of built-up areas. High positive NDBI values indicate the dominance of built-up surfaces, while low or negative values represent non-built-up or natural areas.

By combining the results of both the vegetation index and built-up index, the filtering process identifies and excludes pixels with low vegetation (pixels for which both EVI and NDVI are less than or equal to 0.2) \emph{and} high built-up signatures (pixels that have positive NDBI values). 
\subsection{Random Forest (RF) Classification}
The masking operation described above aims to retain only the non-built-up or natural regions for input to the classification module. For the purpose, we utilize an RF classifier which is an example of ensemble learning where each model is a decision tree. Ensemble learning creates a stronger model by aggregating the predictions of multiple weak models, such as decision trees in our case. To train the RF classifier, we need to have at least two classes. 

We combine multiple sample points along with their corresponding class labels (representing trees or non-trees), divide the samples into an 80\% training set and a 20\% validation set, train a random forest classifier with the features described above, and then use the trained classifier to classify the input image. 
In the process of an RF model training, the user defines the number of features at each node in order to generate a tree. The classification of a new dataset is done by passing down each case of the dataset to each of the grown trees, then the forest chooses a class having the most votes of the trees for that case. More details on RF can be found in Breiman~\cite{breiman2001random}. The main motivation behind choosing RF for this study is its ability to efficiently handle large and high dimensional datasets ~\cite{diaz2006gene, archer2008empirical}.

\section{Evaluation}
The proposed methodology is applied to satellite imagery for the year 2021 and its performance compared to other benchmarks is shown in Table \ref{tab:eval1}. As the results indicate, RF with all multi-spectral bands performs better than the ESA WorldCover product~\cite{van2021esa} and DeepLabv3~\cite{chen2017deeplab}. RF with spectral indices achieve higher accuracy as compared to RF with only spectral bands. Finally, the proposed model accomplishes higher pixel-wise accuracy and Kappa score as compared to all other models (see Table~\ref{tab:eval2}).

Results indicating the effect of feature selection with the proposed methodology are provided in Table~\ref{tab:eval2}. Clearly, as the feature selection set increases, the pixel-wise accuracy and Kappa score increases. It implies pixel-wise accuracy is directly proportional to our feature selection. We choose the Kappa coefficient as a performance metric because it represents the extent to which classes on the ground are correct representations of the classes on the map.

Finally, qualitative results are illustrated in Fig.~\ref{fig:qualtitative}. The ground truth tree cover of LUMS study region is $23$ acres while the predicted area using the proposed model is $25$ acres. It is important to note that our proposed model operating on low resolution imagery with suitable feature selection and masking operations performs better than a DeepLabv3~\cite{chen2017deeplab} deep learning architecture trained on high resolution imagery despite the computational complexity of the former being extremely low compared to the latter.

\section{Conclusion}
The paper proposes a methodology for estimating urban tree cover using RF classification with appropriately selected multispectral features and masking. The proposed methodology exhibits superior performance compared to classical RF classifiers that solely utilize spectral bands, as well as surpassing state-of-the-art models such as the European Space Agency (ESA) WorldCover $10$m $2020$ product~\cite{van2021esa} and DeepLabv3~\cite{chen2017deeplab} deep learning architecture trained on high resolution imagery. Our future work aims to apply the proposed technique to estimate tree cover across entire cities in Pakistan.



\bibliographystyle{IEEEbib}
\bibliography{refs}

\end{document}